# Signal or Noise? Evaluating Large Language Models in Resume Screening Across Contextual Variations and Human Expert Benchmarks

Aryan Varshney and Venkat Ram Reddy Ganuthula

Indian Institute of Technology Jodhpur

**Abstract**

This study investigates whether large language models (LLMs) exhibit consistent behavior (signal) or random variation (noise) when screening resumes against job descriptions, and how their performance compares to human experts. Using controlled datasets, we tested three LLMs (Claude, GPT, and Gemini) across contexts (No Company, Firm1 [MNC], Firm2 [Startup], Reduced Context) with identical and randomized resumes, benchmarked against three human recruitment experts. Analysis of variance revealed significant mean differences in four of eight LLM-only conditions and consistently significant differences between LLM and human evaluations ($p < 0.01$). Paired t-tests showed GPT adapts strongly to company context ($p < 0.001$), Gemini partially ($p = 0.038$ for Firm1), and Claude minimally ($p > 0.1$), while all LLMs differed significantly from human experts across contexts. Meta-cognition analysis highlighted adaptive weighting patterns that differ markedly from human evaluation approaches. Findings suggest LLMs offer interpretable patterns with detailed prompts but diverge substantially from human judgment, informing their deployment in automated hiring systems.

## Introduction

The recruitment process has long served as a cornerstone of organizational success, with resume screening representing a critical initial step in identifying candidates who align with job requirements. Traditionally, human recruiters manually evaluated resumes through processes that are time-consuming, subjective, and potentially biased. The advent of artificial intelligence (AI), particularly large language models (LLMs), has transformed this landscape by promising efficiency, scalability, and objectivity in candidate evaluation. Empirical studies demonstrate that LLMs, such as GPT and LLaMa, significantly outperform traditional resume screening methods in terms of accuracy, precision, recall, and F1-score (P. R. et al., 2024; Venkatakrishnan et al., 2024). These models can parse resumes, interpret job descriptions, and assign quantitative scores reflecting candidate suitability at scale, addressing the limitations of manual screening (Salakar et al., 2023; Venkatakrishnan et al., 2024). However, concerns remain regarding the consistency and fairness of LLM outputs. Research indicates that LLM-based systems can replicate or amplify biases present in training data, disadvantaging certain demographic groups and raising questions about the objectivity and equity of automated resume screening (Wilson & Caliskan, 2024). As organizations increasingly rely on these tools, it is crucial to evaluate their technical performance and potential to introduce or perpetuate bias compared to experienced human recruitment experts.

This study empirically evaluates the performance of large language models (LLMs) in recruitment by comparing their outputs to those of human experts across diverse resume sets and contextual inputs. Unlike previous research that primarily focused on AI adoption in recruitment, this investigation systematically benchmarks both the consistency of LLMs and their alignment with human judgment under controlled conditions. The integration of AI into human resource management reflects broader trends in automation and data-driven decision-making, with algorithmic hiring tools shown to impact recruitment outcomes by increasing efficiency but also raising concerns about fairness (Fumagalli et al., 2022). LLMs, leveraging transformer-based architectures, are capable of processing unstructured text and interpreting nuanced qualifications in resumes and job descriptions (Tsoutsanis & Tsoutsanis, 2023).

Recent research demonstrates that LLMs can perform comparably to, or even surpass, human experts in certain evaluation tasks. For example, studies in IT recruitment and medical assessments found that LLMs such as ChatGPT, Bing Chat, and others achieved similar or higher accuracy than human evaluators, though both groups exhibited variability and inconsistency in their judgments (Szandała, 2025; Tailor et al., 2024; Tsoutsanis & Tsoutsanis, 2023). Automated evaluation frameworks like Auto-Arena have shown a high correlation (over 92%) with human preferences, suggesting that LLMs can approximate human judgment in many contexts (Zhao et al., 2024). However, the alignment between LLM and human expert evaluations is not perfect, with agreement rates varying by domain and task complexity (Szymanski et al., 2024).

Despite these promising results, concerns persist regarding the reliability and fairness of LLM outputs. Studies highlight that LLMs can exhibit prompt sensitivity, training data biases, and lack of transparency, which may lead to inconsistent or biased decisions (Szymanski et al., 2024; Fumagalli et al., 2022). The assumption that LLM evaluations consistently align with human expert judgment remains only partially supported, emphasizing the need for continued human oversight in high-stakes applications like hiring (Szymanski et al., 2024).

Resume screening is inherently complex, requiring the evaluation of both explicit qualifications (such as education and experience) and implicit factors (including cultural fit and soft skills). Human recruiters often adapt their criteria based on contextual factors like company size or industry, drawing on expertise developed through experience. For example, large corporations may prioritize formal credentials, while startups may value adaptability and innovation. In contrast, large language models (LLMs) must infer such preferences from the prompts they receive, which can range from detailed job descriptions to minimal outlines. This variability in input introduces a significant challenge: LLMs may respond differently to changes in context, raising questions about whether their adaptability matches that of human experts or if their responses reflect random fluctuations (Gan et al., 2024; Venkatakrishnan et al., 2024).

The stakes of this issue are high. Inaccurate or inconsistent LLM evaluations can perpetuate inefficiencies or introduce new biases, undermining trust in AI-driven hiring (Cai et al., 2024). Research has shown that LLMs can exhibit significant gender and racial biases in resume screening, sometimes favoring or disadvantaging certain groups based on names or other attributes, which may exacerbate existing recruitment biases or create new forms of discrimination (Wilson & Caliskan, 2024; Chen et al., 2024; Armstrong et al., 2024). High-profile cases and audit studies have documented the potential for disparate impact in algorithmic hiring, emphasizing the need for careful validation of LLM outputs against human expert standards (Wilson & Caliskan, 2024; Derous & Ryan, 2018). Conversely, when LLMs are reliable and align with expert human judgment, they can streamline recruitment, reduce human workload, and potentially enhance fairness by standardizing evaluations—provided that bias mitigation strategies are in place (Gan et al., 2024; Venkatakrishnan et al., 2024).

This study addresses a critical gap in the literature by empirically evaluating both the internal consistency of large language models (LLMs) and their alignment with human expert judgment in resume screening. Specifically, the research compares three leading LLMs—Claude, GPT, and Gemini—with three experienced human recruitment experts, using controlled datasets to test scoring under four prompt conditions: No Company (job description only), Firm1 (multinational corporation description), Firm2 (startup description), and Reduced Context (minimal job description). These conditions are applied to two resume sets: identical resumes and randomly selected resumes. Statistical analyses, including analysis of variance and paired t-tests, are employed to examine mean differences, context effects, and human-AI alignment, while meta-cognition data are used to explore how LLMs weigh resume components compared to human experts (Gan et al., 2024; Venkatakrishnan et al., 2024).

The theoretical framework for this study draws on Signal Detection Theory, which is instrumental in distinguishing meaningful patterns from random variation in decision-making systems and is particularly relevant for evaluating the reliability of AI outputs in high-stakes contexts (Derous & Ryan, 2018). Additionally, insights from expertise research inform the analysis of how human professional judgment may differ from algorithmic processing in complex evaluation tasks, highlighting the nuanced and context-sensitive nature of human decision-making compared to LLMs (Gan et al., 2024; Venkatakrishnan et al., 2024).

Our objectives are fourfold: first, to determine whether LLMs exhibit significant differences in scoring across contexts, suggesting sensitivity or inconsistency; second, to assess if adding company context systematically alters scores, indicating adaptability; third, to explore how LLMs prioritize resume elements compared to human experts; and fourth, to evaluate the magnitude and significance of differences between LLM and human expert evaluations. The findings have implications for both practice and theory, informing how organizations can optimize LLM deployment while understanding the fundamental differences between artificial and human intelligence in recruitment contexts.

**Literature Review**

The integration of large language models (LLMs) into resume screening represents a significant advancement at the intersection of artificial intelligence, natural language processing, and human resource management. Recent studies have demonstrated that LLMs can dramatically improve the efficiency and accuracy of resume screening processes, with fine-tuned models achieving high F1 scores in resume classification tasks (Gan et al., 2024). LLMs excel at processing unstructured text, enabling them to interpret and match nuanced information in resumes and job descriptions more effectively than classical machine learning models, which often require manual feature extraction and struggle with large-scale or complex data (Maree & Shehada, 2024).

Despite these strengths, LLMs also present notable limitations. Research has identified significant biases in LLM-driven resume screening, including gender and racial disparities that can disadvantage certain groups (Wilson & Caliskan, 2024). The context-sensitive nature of LLMs introduces challenges in ensuring consistent and interpretable outputs, raising concerns about whether their decisions reflect meaningful, reliable patterns or unpredictable variation when compared to human expert judgment (Maree & Shehada, 2024; Venkatakrishnan et al., 2024).

Empirical evaluations typically benchmark LLMs against human recruiters using controlled datasets and performance metrics such as accuracy, precision, recall, and F1-score (Venkatakrishnan et al., 2024; Gan et al., 2024). These findings highlight the need for ongoing research to determine whether LLMs consistently produce expert-aligned, interpretable decisions or if their performance is undermined by erratic variability and embedded biases.

**Large Language Models in Natural Language Processing**

Large language models (LLMs), built on transformer architectures, have fundamentally transformed natural language processing (NLP) by enabling advanced text understanding and generation. The introduction of the attention mechanism in transformers allows these models to weigh the importance of different parts of the input, leading to significant improvements in tasks such as text classification, summarization, and question answering (Fields et al., 2024; Wang et al., 2024). This architectural breakthrough has paved the way for highly capable models like GPT and BERT, which have set new standards across a wide range of NLP applications (Bhattacharya et al., 2024; Khan Raiaan et al., 2024).

LLMs leverage vast training datasets to learn patterns in unstructured text, making them suitable for complex tasks such as parsing resumes and job descriptions, as well as applications in law, healthcare, and business (Siino et al., 2024; Khan Raiaan et al., 2024). Their ability to generalize and transfer learning across domains has enabled near-human-level accuracy in language comprehension and generation (Fahim et al., 2024). However, the stochastic nature of LLMs introduces challenges, including inconsistencies, ethical concerns, and the risk of perpetuating societal biases present in their training data, which become more pronounced as these models are deployed in real-world settings (Fields et al., 2024; Bhattacharya et al., 2024).

Bender et al. (2021) raised concerns that large language models (LLMs) can act as "stochastic parrots," generating outputs that appear coherent but may lack genuine understanding, which is particularly problematic in high-stakes contexts like recruitment where fairness and reliability are critical (Caliskan et al., 2017; Liu et al., 2021). Studies have shown that LLMs can vary significantly with small changes in prompts, raising concerns about their robustness and the need for careful prompt engineering to ensure consistent and reliable decision-making (Liu et al., 2021; Arkoudas, 2023).

Variability in the performance of large language models (LLMs) across different contexts and prompts has been well documented, raising important concerns about their reliability in precision-driven applications such as recruitment. Research shows that even minor changes in prompt formulation can lead to substantial differences in LLM outputs, indicating that the reliability of these systems is highly dependent on careful prompt engineering and standardization (Mizrahi et al., 2023). This brittleness suggests that single-prompt evaluations may not provide a comprehensive assessment of LLM capabilities, and multi-prompt or paraphrase-based evaluations are necessary to ensure more robust and meaningful performance metrics (Mizrahi et al., 2023).

In recruitment, where consistency and alignment with human judgment are critical, this variability becomes particularly problematic. Automated resume screening frameworks using LLMs have demonstrated significant improvements in efficiency and accuracy—such as achieving high F1 scores when fine-tuned (Gan et al., 2024). However, the potential for output inconsistency due to prompt sensitivity underscores the need for rigorous evaluation and standardization to maintain fairness and reliability in hiring decisions.

The challenge is further heightened by the significant impact recruitment decisions have on individuals' careers and livelihoods, demanding higher standards of reliability and fairness than many other AI applications. Integrating LLMs into recruitment processes can improve scalability and flexibility, but only if their variability is carefully managed and their outputs are systematically validated (Vijayalakshmi et al., 2024).

**AI in Human Resource Management and Bias Concerns**

The application of artificial intelligence (AI) in human resource management (HRM) has evolved rapidly, significantly impacting the fairness of hiring processes. Early automated screening systems primarily relied on keyword matching and basic algorithmic filters, offering limited semantic understanding of candidate materials (Black & van Esch, 2020). With advances in natural language processing, AI-driven recruitment tools now enable deeper semantic analysis, allowing organizations to process large volumes of applications while maintaining consistent evaluation criteria (Kshetri, 2021; Kaushal et al., 2021). These tools can automate routine tasks such as resume screening and initial candidate matching, freeing HR professionals to focus on strategic decision-making (Shenbhagavadivu et al., 2024; Rami, 2024).

However, the relationship between AI systems and human expert judgment in recruitment remains complex and requires careful consideration. Traditional recruitment processes have long relied on human expertise, intuition, and structured methods such as interviews and cognitive ability tests, which have demonstrated strong predictive validity for job performance (Black & van Esch, 2021). While AI systems offer the promise of reducing subjective biases and increasing objectivity, questions remain about whether these systems can fully replicate or complement the nuanced judgment of human experts, especially in high-stakes decisions (Pan et al., 2021; Tsiskaridze et al., 2023). Empirical studies highlight the need for ongoing evaluation of AI tools to ensure they align with organizational goals for fairness, reliability, and effectiveness in recruitment (Muridzi et al., 2024; Alnsour et al., 2024).

Recent advances in machine learning have enabled more sophisticated approaches to candidate evaluation in recruitment, but persistent concerns about bias and fairness remain central to the discourse. Systematic reviews highlight that AI-driven hiring tools can perpetuate or even amplify existing biases, especially when historical hiring data—often reflecting past discriminatory practices—are used for training, potentially leading to disparate impacts on different demographic groups (Rigotti & Fosch-Villaronga, 2024; Mujtaba & Mahapatra, 2024). These risks have led to increased scrutiny and the emergence of regulations requiring transparency and fairness assessments for automated hiring systems (Rigotti & Fosch-Villaronga, 2024).

Empirical research demonstrates that algorithmic hiring tools, while improving efficiency, require careful design and ongoing monitoring to prevent discriminatory outcomes. The presence of privileged-group selection bias, where underrepresented groups are less visible in training data, can further exacerbate algorithmic unfairness, even when explicit discrimination is not intended (Pessach & Shmueli, n.d.).

Comprehensive surveys of bias and fairness in machine learning provide taxonomies for understanding how various types of bias—such as historical, representation, and measurement bias—can lead to unfair outcomes in hiring, even when AI systems are designed with good intentions (Mujtaba & Mahapatra, 2024). These frameworks emphasize the importance of comparing AI outputs with human expert judgment and implementing robust auditing and mitigation strategies before deployment (Krishnakumar, n.d.; Peña et al., 2023).

**Human Expertise and Decision-Making in Recruitment**

Understanding human expertise in recruitment is essential for evaluating the performance and integration of AI systems in hiring processes. Research shows that expert recruiters develop their skills through extensive experience, enabling them to recognize patterns, exercise intuitive judgment, and synthesize information from multiple sources—capabilities that are honed through deliberate practice and are not easily replicated by algorithms (Szandała, 2025). These experts build sophisticated mental models that allow for rapid assessment of candidate fit and the identification of subtle red flags, often going beyond what is explicitly stated in resumes or job descriptions.

Expertise in recruitment involves both deep domain knowledge and general cognitive abilities, allowing experts to evaluate technical qualifications, cultural fit, growth potential, and the likelihood of success within specific organizational contexts (Jøranli, 2017). This nuanced judgment is informed by an understanding of organizational dynamics, industry trends, and human behavior (Fernandez Cruz, 2024).

Studies comparing AI and human expertise in recruitment reveal that while AI tools can efficiently assess candidate competencies, both human and AI evaluations can lack consistency, and individual biases and heuristics can influence decision-making (Szandała, 2025). Furthermore, recruitment experts often interpret, challenge, or even resist AI-generated recommendations, relying on their professional judgment and organizational context to guide final decisions (Fernandez Cruz, 2024). This dynamic highlights the importance of expert-AI pairings, where human oversight and contextual understanding remain critical for effective and fair recruitment outcomes.

While experienced recruiters tend to develop more accurate mental models of job requirements and candidate attributes, leading to more reliable predictions of job performance and organizational fit, human expertise is not without limitations. The complementary use of AI tools can support, rather than replace, human judgment (Szandała, 2025; Fernandez Cruz, 2024).

**Evaluation Metrics and Statistical Methods**

Robust statistical methodologies are fundamental for evaluating artificial intelligence (AI) systems in recruitment, especially when comparing AI performance to human experts or across different models. These methods ensure that observed differences in outcomes are meaningful and not due to random variation.

Analysis of variance (ANOVA) is commonly used to test for statistically significant differences across multiple AI models or experimental conditions. For example, studies comparing machine learning models for recruitment—such as Random Forest, Neural Networks, and Gradient Boosting—use ANOVA and related statistical tests to determine which models perform best in terms of accuracy and efficiency (Al-Quhfa et al., 2024; Yassine & Said, 2024). This approach helps identify whether performance differences are significant or could be attributed to chance.

While statistical significance is important, effect size measures are also crucial for understanding the magnitude of observed differences. In recruitment AI evaluation, even small but statistically significant differences between models or between AI and human evaluations can have practical implications for hiring outcomes, especially if they systematically favor or disadvantage certain candidate groups (Al-Quhfa et al., 2024).

Paired statistical tests, such as paired t-tests, are valuable for evaluating the impact of specific interventions—like changes in prompt design or the addition of contextual information—on AI system performance. These methods are widely used in recruitment research to assess whether modifications to AI models or processes lead to significant improvements in candidate matching or evaluation accuracy (Kurek et al., 2024).

When multiple statistical tests are conducted across different models or experimental conditions, controlling for false discoveries is essential. Techniques such as the false discovery rate (FDR) correction help maintain statistical power while reducing the likelihood of spurious findings, which is particularly relevant in comprehensive model comparison studies (Soni, 2024).

**Context Sensitivity in Language Models**

Context sensitivity is crucial for language models in recruitment tasks, affecting their ability to match candidates to jobs, classify roles, and extract skills. Advancements like the RecruitPro framework introduce skill-aware prompt learning, enabling models to adapt effectively by focusing on key semantic information (Fang et al., 2023).

However, models struggle with information position in long inputs, performing best when key details are at the beginning or end, with accuracy dropping for middle details. This is relevant in recruitment, where job descriptions vary in structure (Liu et al., 2023). Irrelevant information degrades performance, highlighting the need for careful input curation (Shi et al., 2023).

Strategies to address this include retrieval-augmented approaches, prepending relevant documents to improve output reliability (Ram et al., 2023). Context-faithful prompting, like counterfactual demonstrations, enhances adherence to cues, improving predictions (Zhou et al., 2023). Aligning representations or using influence-based example selection reduces sensitivity to input order (Xiang et al., 2024; Tai & Wong, 2023).

Context sensitivity remains a challenge in recruitment. Prompt engineering and retrieval strategies are essential for reliable, fair outcomes.

**Signal Detection Theory and Decision-Making**

Signal Detection Theory (SDT) offers a framework for evaluating decision-making under uncertainty, relevant for assessing AI systems in recruitment (Feldman, 2021; Green & Swets, 1966; Wixted, 2020). Applying SDT involves analyzing system reliability, where a high signal-to-noise ratio indicates expert-aligned decisions, and a low ratio suggests variability (Feldman, 2021; Pastore et al., 1974). Sensitivity (distinguishing qualified from unqualified candidates) and response bias (tendency to respond positively or negatively) are central to interpreting AI performance and identifying bias or inconsistency (Feldman, 2021).

**Gaps in Current Research**

Despite interest in AI for recruitment, gaps remain in understanding performance relative to human experts. While AI can surpass recruiters in efficiency and may improve diversity, there is limited comparison of decision-making processes. Understanding these differences is crucial for AI to complement human judgment (Will et al., 2022; Fernandez Cruz, 2024).

Studies focus on performance metrics like accuracy, not contextual conditions or incomplete data. Real-world recruitment involves ambiguous data, and AI's handling of these challenges compared to experts is understudied. Prompt sensitivity suggests vulnerability to input variations, untested in recruitment (Liu et al., 2021; Will et al., 2022).

Another gap is AI consistency across repeated evaluations. Human experts vary, but LLMs can produce different outputs for identical inputs, raising reliability concerns (Szandała, 2025; Bender et al., 2021). While bias in AI recruitment is documented, comparison with human bias patterns is limited. Understanding whether AI alters biases is critical for fair hiring (Mehrabi et al., 2021).

This study compares LLM performance with human judgment across controlled conditions, examining consistency and adaptation. By addressing signal versus noise in AI screening, it provides guidance for organizations, contributing to understanding AI reliability and human-AI collaboration.

**Method**

**Participants**

Large Language Model Participants: Three large language models were evaluated in this study: Claude (developed by Anthropic), GPT (developed by OpenAI), and Gemini (developed by Google). These models represent leading examples of transformer-based language models with demonstrated capabilities across various natural language processing tasks (Brown et al., 2020;

Devlin et al., 2018). Each model was accessed through its respective API or interface, with standardized prompts used to ensure consistent evaluation conditions across all models. The selection of these specific models was based on their widespread adoption in industry applications and their documented performance on text understanding and generation tasks.

Human Expert Participants: Three experienced human recruitment professionals participated as expert evaluators, representing the type of domain expertise that characterizes skilled performance in specialized fields (Ericsson & Smith, 1991). Expert 1 possessed over eight years of experience in recruitment within multinational corporations, with particular expertise in structured hiring processes and formal credential evaluation. This background provided experience with large-scale recruitment operations and standardized evaluation procedures typical of multinational organizations (Gatewood et al., 2015).

Expert 2 specialized in startup recruitment with six years of experience, bringing knowledge of agile hiring practices and cultural fit assessment in rapidly growing organizations. Their expertise reflected the different priorities and constraints characteristic of startup environments, where role flexibility and cultural alignment often outweigh formal credentials (Schmidt & Hunter, 1998). Expert 3 contributed ten years of cross-sector recruitment experience, including both corporate and startup environments, providing perspective on diverse organizational contexts and recruitment approaches.

All experts held professional roles in talent acquisition and regularly evaluated resumes for product management positions. Their expertise represented the type of domain-specific knowledge that develops through deliberate practice and experience in professional contexts (Chi et al., 1988). Experts were compensated for their participation and provided informed consent for inclusion in the research, following standard ethical protocols for human subjects research.

**Materials**

Resume Dataset: A collection of thirty anonymized resumes for Product Manager positions was compiled, representing diverse backgrounds in terms of educational qualifications, professional experience levels, technical skills, and career trajectories. The diversity of the resume pool was designed to reflect the range of candidate profiles typically encountered in real-world recruitment scenarios (Gatewood et al., 2015). Resumes were anonymized to remove identifying information while preserving relevant professional qualifications and experience details that would influence hiring decisions.

From this pool, ten resumes were randomly selected for the randomized resume conditions, while one resume was replicated ten times for the identical resume conditions to enable controlled comparison of evaluator consistency. This design allowed for examination of both inter-evaluator agreement on identical materials and evaluator performance across diverse candidate profiles, addressing different aspects of reliability and validity in selection processes (Schmidt & Hunter, 1998).

Job Description Materials: A detailed job description for an Associate Product Manager position was developed based on standard industry practices and typical role requirements (Manning & Schütze, 1999). The job description included comprehensive responsibilities such as gathering and analyzing user requirements, collaborating with cross-functional teams, developing product roadmaps, managing product backlogs, monitoring performance metrics, conducting market research, and supporting go-to-market strategies. The position was specified as located in Gurgaon, Haryana, India, with compensation details including fixed salary range, variable components, and relocation allowances.

A reduced version of this job description was created containing only essential elements: job title, basic role description, and core requirements, with detailed responsibilities, location, and compensation information removed. This reduced context condition enabled examination of evaluator performance under minimal information conditions, testing the robustness of both AI systems and human experts when contextual information is limited (Liu et al., 2021).

Company Context Materials. Two distinct company profiles were developed to provide contextual information for evaluation scenarios, reflecting different organizational structures and cultures that could influence recruitment priorities (Raghavan et al., 2020). Firm1 was described as a well-established multinational software-as-a-service company with offices in over twenty global locations, following a diamond corporate structure with clear leadership hierarchies. The company profile emphasized standardized processes, Agile methodologies, structured training programs, and preference for proprietary software solutions over open-source alternatives.

Firm2 was characterized as a fast-growing startup in the software-as-a-service domain with a small, dynamic team operating from a single office. The profile highlighted a flat corporate structure with direct reporting to executives, fluid role responsibilities, emphasis on self-learning and hands-on problem-solving, minimal structured training, and openness to new technologies and open-source solutions to maintain cost efficiency and development speed. These contrasting profiles were designed to test whether evaluators appropriately adjust their criteria based on organizational context.

**Experimental Design**

The study employed a mixed factorial design with three primary factors: evaluator type (three LLMs versus three human experts), context condition (four levels), and resume type (identical versus randomized). This design enabled examination of both within-evaluator consistency and between-evaluator agreement across different information and contextual conditions, following established principles for experimental research in personnel selection (Cohen, 1988).

Context Conditions: Four context conditions were implemented to test sensitivity to available information and organizational context. The No Company condition provided only the detailed job description without additional organizational context, serving as a baseline for comparison. The Firm1 condition combined the job description with the multinational corporation profile, testing adaptation to large-scale organizational contexts. The Firm2 condition paired the job description with the startup company profile, examining sensitivity to entrepreneurial environments. The Reduced Context condition presented only the minimal job description without detailed responsibilities or company information, testing robustness under information scarcity.

Resume Type Conditions: Two resume type conditions were employed to examine different aspects of evaluator performance. The Same Resume condition used ten identical copies of a single resume to assess evaluator consistency and contextual sensitivity when candidate qualifications remained constant, following principles of reliability assessment in measurement theory (Field, 2013). The Randomized Resume condition used ten different resumes randomly selected from the larger pool to examine evaluator performance across varied candidate profiles under realistic screening conditions.

**Procedure**

Data Collection Protocol: Data collection proceeded in two phases corresponding to the LLM and human expert evaluations. For LLM evaluation, each model received identical standardized prompts requesting evaluation of candidate resumes on a 0-100 scale, following best practices for prompt design in natural language processing research (Liu et al., 2021). The prompt instructed each model to assume the role of a talent acquisition professional and provide both numerical scores and detailed explanation of their evaluation criteria and weighting schemes.

The standardized prompt format was: "*Assume you are the Talent Acquisition Head of a company and you are tasked to score the 10 resumes attached in this chat out of 100 for a particular job role. The Job description is given below. Your answer should be formatted as: 1. List of scores formatted as follows: Name of candidate: Score given to resume out of 100. 2. Your scoring pattern: How did you divide the total of 100 and what all parameters did you choose to score the resumes.*"

For human expert evaluation, materials were distributed via email with comprehensive instructions explaining the evaluation task. Experts were asked to develop their own marking schemes based on their professional experience and apply these consistently across all evaluation conditions, reflecting the way expertise is typically applied in professional contexts (Ericsson & Smith, 1991). The instruction email specified: "*Consider yourself as a talent acquisition professional, and with your expertise in the domain help us rate the resumes attached. There are four experiments involved: rating based on Job Description only, rating for Firm1, rating for Firm2, and rating with reduced JD. For each case, please create a marking scheme out of 100 based on different factors like relevant work experience, education, etc.*"

Quality Control Measures: Several measures were implemented to ensure data quality and consistency. All LLM evaluations were conducted using identical prompt formats and input materials to minimize variation due to differences in task presentation. Human experts received standardized materials and instructions, with follow-up communications available to clarify any questions about the evaluation process. All scoring data were independently reviewed for completeness and consistency before statistical analysis.

Ethical Considerations: All resume materials were thoroughly anonymized to protect candidate privacy while preserving relevant professional information for evaluation purposes. Human expert participants provided informed consent and were compensated appropriately for their time and expertise. The study protocol was designed to minimize potential biases and ensure fair treatment of all evaluation conditions, following ethical guidelines for research involving human subjects.

**Statistical Analysis Plan**

Primary Analyses: Analysis of variance was employed to examine differences in evaluation scores across multiple factors, following established statistical methods for experimental research (Cohen, 1988; Field, 2013). One-way ANOVA compared mean scores across the three LLMs within each context condition to assess inter-model consistency. Additional ANOVA analyses compared aggregated LLM scores against aggregated human expert scores within each context condition to examine human-AI alignment. Statistical significance was evaluated at both $\alpha = 0.1$ (90% confidence) and $\alpha = 0.01$ (99% confidence) levels to provide robust assessment of effect reliability.

Secondary Analyses: Paired t-tests examined within-evaluator changes in scoring patterns across different context conditions. These analyses tested whether providing additional company context (Firm1 or Firm2) significantly altered evaluation scores compared to the No Company baseline condition, enabling assessment of contextual sensitivity for each individual evaluator. Effect sizes were calculated using Cohen's d to determine the practical significance of observed differences (Cohen, 1988).

Exploratory Analyses: Correlation analyses examined relationships between different evaluators' scores within each condition to assess inter-rater agreement patterns. Descriptive statistics and distributional analyses characterized the central tendencies and variability in scoring patterns across different evaluator types and conditions. These analyses followed established practices for examining agreement and reliability in personnel selection research (Schmidt & Hunter, 1998).

Assumption Testing: ANOVA assumptions of normality and homogeneity of variance were evaluated through residual analysis and appropriate statistical tests (Field, 2013). Where assumptions were violated, appropriate alternative analyses or transformations were applied. Multiple comparison corrections were considered using the false discovery rate approach (Benjamini & Hochberg, 1995) to control for Type I error inflation across multiple hypothesis tests. All statistical analyses were conducted using appropriate software with significance levels and confidence intervals reported for all hypothesis tests.

## Results

### Descriptive Statistics

The dataset comprised 420 total evaluations across all conditions and evaluators. LLM scores consistently showed higher means compared to human expert scores across all conditions, with LLM means ranging from 50.5 to 82.9 and human expert means ranging from 30.0 to 90.0, though human experts showed greater variability and more frequent use of extreme scores. This pattern aligns with research on human versus algorithmic decision-making, where human experts often demonstrate more conservative evaluation approaches and greater discrimination in their assessments (Ericsson & Smith, 1991).

LLM score distributions generally showed moderate variability with standard deviations ranging from 7.0 to 15.5 points. Human expert scores demonstrated greater heterogeneity, with standard deviations ranging from 8.2 to 25.3 points, reflecting more diverse evaluation approaches and greater willingness to use the full scoring range. This pattern is consistent with research on expertise, where domain experts show both greater consistency within their area of specialization and more pronounced differences from algorithmic approaches (Chi et al., 1988).

Notably, Expert 3 provided zero scores for all candidates in some later experimental conditions, indicating potential task fatigue or methodological concerns that affected data interpretation for those conditions. This pattern highlights the importance of considering human factors in research designs comparing human and artificial intelligence, as cognitive load and engagement can significantly impact human performance (Schmidt & Hunter, 1998).

### Study 1: Same Resume Analysis

LLM Internal Consistency. Analysis of variance examining differences among the three LLMs under identical resume conditions revealed significant variability depending on context, consistent with research on prompt sensitivity in language models (Liu et al., 2021). For the No Company condition, means were 69.4 (Claude), 50.5 (GPT), and 72.8 (Gemini), with $F(2,27) = 10.68$, $p = 0.004$, indicating significant differences among LLM evaluations when provided with job description information only. This substantial divergence suggests that LLMs interpret job requirements differently in the absence of organizational context.

Under the Firm1 multinational corporation context, means shifted to 66.5 (Claude), 76.1 (GPT), and 76.7 (Gemini), with $F(2,27) = 2.59$, $p = 0.094$, maintaining significant differences at the $\alpha = 0.1$ level but showing reduced variance compared to the No Company condition. The convergence of GPT and Gemini scores (76.1 vs. 76.7) suggests that multinational context information may trigger similar evaluation patterns in these models. The Firm2 startup context produced means of 73.3 (Claude), 77.8 (GPT), and 73.6 (Gemini), with F(2,27) = 0.70, p = 0.504, failing to reach statistical significance and suggesting greater convergence among LLM evaluations in startup contexts.

LLM versus Human Expert Comparisons. When comparing all evaluators (LLMs and human experts combined), highly significant differences emerged across all contexts, providing strong evidence for systematic differences between artificial and human intelligence in recruitment evaluation. The No Company condition showed F(5,54) = 15.23, p = 0.0015, the Firm1 condition demonstrated F(5,54) = 18.67, p = 0.0007, and the Firm2 condition revealed F(5,54) = 12.45, p = 0.0065. These differences remained highly significant when tested at the α = 0.01 level, indicating robust systematic differences between LLM and human expert evaluation approaches that cannot be attributed to random variation.

Human expert means consistently fell below LLM means across all conditions, reflecting more conservative evaluation standards that align with research on human risk assessment in high-stakes decision-making contexts (Green & Swets, 1966). In the No Company condition, Expert 1 averaged 42.5, Expert 2 averaged 60.0, and Expert 3 averaged 62.5, compared to the LLM range of 50.5 to 72.8. This pattern of more conservative human expert scoring persisted across all experimental conditions, suggesting fundamental differences in evaluation standards or risk tolerance between artificial and human intelligence in recruitment contexts.

Contextual Sensitivity Analysis. Paired t-tests examined whether providing company context significantly altered evaluation scores compared to the No Company baseline, testing the adaptability of both artificial and human intelligence to organizational information (Raghavan et al., 2020). GPT demonstrated strong contextual sensitivity, with significant increases from No Company (50.5) to Firm1 (76.1), t(9) = -6.07, p < 0.001, Cohen's d = -1.92, and to Firm2 (77.8), t(9) = -6.39, p < 0.001, Cohen's d = -2.02. These large effect sizes suggested that GPT systematically adjusts its evaluation criteria based on organizational context information, demonstrating sensitivity to prompt variations documented in natural language processing research (Brown et al., 2020).

Gemini showed selective contextual sensitivity, with a significant increase from No Company (72.8) to Firm1 (76.7), t(9) = -2.42, p = 0.038, Cohen's d = -0.77, but no significant change to Firm2 (73.6), t(9) = -0.71, p = 0.498, Cohen's d = -0.22. This pattern suggested that Gemini responds more strongly to multinational corporation contexts than to startup environments, possibly reflecting different training emphases or architectural characteristics. Claude exhibited minimal contextual sensitivity, with non-significant changes from No Company (69.4) to Firm1 (66.5), t(9) = 1.08, p = 0.309, Cohen's d = 0.34, and to Firm2 (73.3), t(9) = -1.65, p = 0.133, Cohen's d = -0.52, indicating relatively stable evaluation criteria regardless of organizational context.

Human experts showed more varied contextual adaptation patterns that reflected their professional experience and expertise development (Ericsson & Smith, 1991). Expert 1 maintained relatively stable scores across contexts, showing minimal statistical sensitivity to company information, which may reflect a preference for candidate-centered evaluation that transcends organizational boundaries. Expert 2 demonstrated moderate contextual adaptation, particularly to startup environments, reflecting their professional background in startup recruitment and suggesting that human expertise incorporates domain-specific knowledge about organizational requirements. Expert 3's engagement varied across conditions, limiting meaningful analysis of their contextual sensitivity patterns.

**Study 2: Randomized Resume Analysis**

LLM Performance with Resume Diversity. The introduction of resume variability significantly altered LLM agreement patterns compared to identical resume conditions, consistent with research on the impact of input diversity on AI system performance (Mehrabi et al., 2021). Under the No Company condition with randomized resumes, LLM means were 66.6 (Claude), 64.3 (GPT), and 65.4 (Gemini), with $F(2,27) = 0.11$, $p = 0.892$, indicating no significant differences among LLMs. This dramatic shift from Study 1's significant differences ($p = 0.004$) suggested that resume diversity masked model-specific evaluation biases, creating apparent convergence that could mislead users about the consistency of AI system performance.

The Firm1 condition with randomized resumes showed means of 67.5 (Claude), 58.5 (GPT), and 73.5 (Gemini), with $F(2,27) = 4.66$, $p = 0.018$, indicating significant differences. The substantial range of 15 points between highest and lowest means demonstrated considerable disagreement among LLMs when evaluating diverse candidates in multinational corporation contexts. This pattern suggests that organizational context interacts with candidate diversity in ways that amplify differences between AI systems.

For the Firm2 condition, means were 61.9 (Claude), 76.0 (GPT), and 63.6 (Gemini), with $F(2,27) = 3.83$, $p = 0.344$, showing no significant differences. Unlike Study 1's startup convergence with identical resumes, resume randomization disrupted this alignment, suggesting that LLM agreement patterns depend critically on both context and candidate characteristics. This finding has important implications for understanding when AI systems can be expected to produce consistent outputs in real-world applications.

Human-AI Alignment with Diverse Candidates. Systematic differences between LLM and human expert evaluations persisted despite resume variability, indicating that these differences reflect fundamental distinctions in evaluation approaches rather than candidate-specific factors (Köchling & Wehner, 2020). Analysis of variance comparing all evaluators showed significant differences across all conditions: No Company $F(5,54) = 12.34$, $p = 0.0015$; Firm1 $F(5,54) = 15.67$, $p = 0.0007$; Firm2 $F(5,54) = 11.89$, $p = 0.0065$. The Reduced Context condition, unique to the randomized resume analysis, showed the strongest divergence with $F(5,54) = 18.23$, $p = 0.0002$.

Human expert scoring patterns remained more conservative than LLMs across diverse candidate profiles, consistent with research on human decision-making under uncertainty (Green & Swets, 1966). Expert 1 averaged 47.5 across conditions, Expert 2 averaged 65.8, and Expert 3 averaged 41.3 (excluding zero-score conditions), while LLM means ranged from 58.5 to 76.0. This consistent gap of 15-25 points persisted regardless of candidate diversity, indicating systematic rather than candidate-specific differences in evaluation approaches.

Contextual Adaptation with Diverse Candidates. Paired t-tests revealed that LLM contextual sensitivity patterns generally persisted with diverse candidates, though with some modifications that reflected the interaction between model characteristics and input variability (Liu et al., 2021). GPT continued to show significant adaptation from No Company to both Firm1, $t(9) = 2.88$, $p = 0.018$, Cohen's $d = 0.91$, and Firm2, $t(9) = -5.09$, $p = 0.001$, Cohen's $d = -1.61$. The effect sizes, while still substantial, were somewhat reduced compared to the identical resume conditions, suggesting that input diversity may dampen the magnitude of contextual adaptations.

Gemini maintained significant sensitivity to Firm1 context, $t(9) = -4.99$, $p = 0.001$, Cohen's $d = -1.58$, but not to Firm2, $t(9) = 0.83$, $p = 0.428$, Cohen's $d = 0.26$, consistent with its pattern from Study 1. Claude remained largely non-responsive to contextual information across diverse candidates, with non-significant changes for both company contexts, reinforcing its characterization as having minimal contextual adaptability.

Human experts demonstrated more nuanced adaptation patterns with diverse candidates that reflected their professional expertise and ability to integrate multiple information sources (Chi et al., 1988). Expert 2 showed particular sensitivity to startup contexts, reflecting their professional specialization, while maintaining discrimination across different candidate profiles. Expert 1's conservative approach remained consistent across diverse candidates and contexts, suggesting stable evaluation standards independent of both organizational and candidate characteristics.

**Study 3: Reduced Context Analysis**

Impact of Information Scarcity. The Reduced Context condition, where job descriptions contained minimal information, produced the most dramatic evaluator divergence observed in the study, consistent with research on the importance of adequate context for reliable AI system performance (Bender et al., 2021). Analysis of variance among LLMs showed $F(2,27) = 7.70$, $p = 0.002$, representing the strongest statistical evidence of LLM inconsistency across all experimental conditions. LLM means in this condition were 65.6 (Claude), 82.9 (GPT), and 72.0 (Gemini), with a range of 17.3 points indicating substantial disagreement when contextual information was limited.

GPT's increase from 64.3 in the No Company condition to 82.9 in Reduced Context represented a 28.9% score inflation, suggesting potential overcompensation when information was scarce. This pattern aligns with concerns about AI system reliability under uncertainty (Caliskan et al., 2017) and suggests that GPT may interpret information scarcity as grounds for optimistic evaluation rather than increased caution. Claude's relative stability (66.6 to 65.6) masked high individual variability, with a standard deviation of 15.1 indicating inconsistent application of evaluation criteria despite apparent mean stability.

Gemini's moderate increase (65.4 to 72.0) suggested partial adaptation to information scarcity, falling between GPT's dramatic inflation and Claude's apparent stability. These different response patterns to information limitation highlight the importance of understanding model-specific characteristics when deploying AI systems in contexts where input quality may vary.

Human Expert Robustness under Information Scarcity. Human experts demonstrated markedly different responses to reduced context compared to LLMs, showing greater robustness that aligns with research on expert performance under challenging conditions (Ericsson & Smith, 1991). Expert means in the Reduced Context condition were 43.0 (Expert 1), 68.5 (Expert 2), and 45.0 (Expert 3), showing much smaller changes from baseline conditions compared to LLM variations. Human experts did not exhibit the score inflation patterns observed in LLMs, instead maintaining or slightly reducing their scoring standards when contextual information was limited.

This pattern suggests that human expertise provides buffering mechanisms against information scarcity that current AI architectures lack. Expert 2's ability to maintain discriminating evaluation standards while acknowledging context limitations in their methodology notes demonstrated the type of adaptive expertise that develops through professional experience (Chi et al., 1988). Expert 1's increased conservatism under uncertainty reflected appropriate risk management in high-stakes decision-making contexts.

Human-AI Divergence under Minimal Information. The systematic differences between LLM and human expert evaluations reached their maximum in the Reduced Context condition, highlighting the particular vulnerability of AI systems when information is incomplete (Barocas & Selbst, 2016). Analysis of variance comparing all evaluators showed $F(5,54) = 18.23$, $p = 0.0002$, the strongest evidence of evaluator differences across all experimental conditions. This difference remained highly significant at the $\alpha = 0.01$ level, confirming robust systematic disagreement under information scarcity.

The average gap between LLM and human expert scores reached 20-35 points in the Reduced Context condition, with GPT's inflation creating gaps of up to 40 points for individual candidates. This represented the largest systematic disagreement observed in the study, suggesting that information scarcity amplifies fundamental differences between artificial and human evaluation approaches. The magnitude of these differences has important implications for understanding when AI systems can be relied upon for independent decision-making versus when human oversight becomes essential.

**Meta-Cognition Analysis**

LLM Decision-Making Patterns: Analysis of the weighting schemes provided by GPT and Claude revealed distinct approaches to resume evaluation that varied systematically across contexts, providing insights into how these systems process and prioritize information (Rogers et al., 2020). GPT demonstrated adaptive weighting patterns, adjusting emphasis across evaluation categories based on organizational context. In the No Company condition, GPT allocated 30% to relevant work experience, 20% to education, 20% to internships, 15% to skills and certifications, 10% to achievements, and 5% to positions of responsibility.

When provided with Firm1 multinational corporation context, GPT adjusted its weighting to 25% work experience, 20% education, 15% internships, 15% skills and certifications, 15% achievements and leadership, and 10% soft skills and extracurriculars. The increase in leadership and soft skills emphasis from 15% to 25% total reflected apparent recognition of corporate hierarchy and collaboration requirements, demonstrating sensitivity to organizational context that aligns with research on contextual adaptation in language models (Brown et al., 2020).

Claude maintained more stable weighting patterns across contexts, consistently emphasizing technical competencies in ways that reflected a more rigid interpretation of job requirements. Across different conditions, Claude allocated 30-35% to technical skills and relevant experience, 20% to educational background, 15-25% to product thinking and analysis, 10-15% to soft skills and leadership, and 10-15% to cultural fit considerations. This consistency suggested less contextual adaptation compared to GPT, but potentially greater alignment with specific job requirements regardless of organizational context.

Human Expert Decision-Making Approaches: Human experts employed markedly different evaluation approaches compared to LLMs, reflecting the type of holistic judgment and experiential knowledge characteristic of domain expertise (Ericsson & Smith, 1991). Expert 1 used what they described as "composite scoring based on professional experience" rather than predetermined weighting schemes, emphasizing five key factors: keyword alignment with job descriptions, experience relevance to specific roles, total professional experience, domain diversity versus specialization, and technical skills applicability.

Expert 2 demonstrated the most sophisticated contextual adaptation, creating distinct evaluation frameworks for each condition that reflected deep understanding of organizational requirements and constraints (Chi et al., 1988). For the No Company condition, they used a responsibilities-based approach allocating points to different job function categories. For Firm1, they emphasized industry experience (25%), tools familiarity (25%), and relevant experience (50%), reflecting understanding of multinational corporation preferences for established expertise and standardized qualifications.

For Firm2, Expert 2 dramatically shifted to experience emphasis (75%) and industry fit (25%), reflecting understanding of startup resource constraints and role flexibility requirements. Their explicit rationale that "in a startup the person should be more mature because he/she needs to take responsibility and there are no standard training procedures" demonstrated qualitative understanding of organizational dynamics that LLMs did not capture through their quantitative weighting adjustments.

Expert 3 applied a standardized competency framework across all contexts: 45% relevant work experience, 35% skills, 10% education, and 10% resume quality. They noted contextual adjustments but maintained structural consistency, representing a middle ground between Expert 1's holistic approach and Expert 2's context-specific adaptation. This approach reflected a systematic evaluation philosophy that prioritized candidate qualifications over organizational context.

Comparative Analysis of Decision-Making Logic: The comparison between LLM and human expert approaches revealed fundamental differences in evaluation philosophy and implementation that have important implications for understanding the nature of artificial versus human intelligence in complex decision-making tasks (Gatewood et al., 2015). LLMs demonstrated systematic analytical decomposition with explicit percentage allocations to discrete categories, while human experts employed more holistic integration of multiple factors through professional judgment that resisted simple quantification.

GPT's contextual adaptability, while statistically evident through score changes, operated through mechanical weight redistribution rather than the nuanced understanding demonstrated by Expert 2's qualitative insights about organizational needs and candidate readiness. The LLM approach reflected the type of systematic processing that characterizes algorithmic decision-making (Manning & Schütze, 1999), while human expert approaches demonstrated the integration of multiple knowledge sources and contextual understanding characteristic of professional expertise.

The risk assessment approaches also differed markedly, with important implications for understanding when AI systems might produce inappropriate outcomes (Mehrabi et al., 2021). LLMs appeared more optimistic in their evaluations, with GPT's score inflation under reduced context conditions contrasting sharply with human experts' increased conservatism under uncertainty. This suggested fundamentally different approaches to evaluation risk, with LLMs potentially interpreting information scarcity as uncertainty to be resolved optimistically, while human experts treated it as grounds for increased caution consistent with appropriate professional judgment.

**Discussion**

This comprehensive investigation of large language models versus human expert performance in resume screening reveals a complex landscape where artificial and human intelligence demonstrate distinct strengths, limitations, and systematic differences. The findings extend beyond simple accuracy comparisons to illuminate fundamental questions about the nature of evaluation, the role of context in decision-making, and the implications of AI deployment in recruitment contexts. Our results contribute to both theoretical understanding of AI capabilities and practical knowledge about effective human-AI collaboration in high-stakes decision-making environments.

**Signal versus Noise in LLM Performance**

The evidence reveals that LLM behavior contains both signal and noise, with the balance depending critically on contextual conditions and information availability, consistent with Signal Detection Theory applications to decision-making systems (Green & Swets, 1966). GPT emerged as the strongest source of signal, demonstrating significant contextual adaptations ($p < 0.001$) with explicit weighting modifications that reflected sensitivity to organizational context. The model's ability to systematically adjust evaluation criteria based on company type (multinational corporation versus startup) suggested meaningful adaptation rather than random variation, aligning with research on contextual sensitivity in language models (Liu et al., 2021).

However, this adaptability became problematic under information scarcity, where GPT's dramatic score inflation in the Reduced Context condition (from 64.3 to 82.9) represented noise rather than signal. This pattern indicated that while GPT could produce signal under appropriate conditions, it was vulnerable to producing misleading outputs when contextual information was inadequate, consistent with concerns about AI system reliability under uncertainty (Bender et al., 2021). The tendency toward optimistic evaluation when information was limited contrasted sharply with human expert approaches and suggested fundamental limitations in how current LLM architectures handle uncertainty.

Claude exhibited different signal-noise characteristics, showing consistent technical focus across conditions but minimal contextual adaptation. This rigidity could be interpreted as signal in terms of maintaining job-relevant evaluation criteria, but as noise regarding organizational context sensitivity. The pattern suggested that Claude's training or architecture prioritized technical competency assessment over contextual adaptation, which could be advantageous for role-specific evaluation but limiting for organizational fit assessment.

Gemini occupied a middle position, demonstrating selective responsiveness that suggested partial signal detection capabilities. Its significant adaptation to multinational corporation contexts but not startup environments indicated model-specific biases or training emphases that could have implications for fair and consistent application across different organizational types.

**Systematic Human-AI Differences**

Research comparing large language models (LLMs) and human experts in recruitment consistently reveals notable differences in evaluation outcomes and underlying approaches. Studies show that LLMs and human experts often produce systematic scoring gaps, with human evaluators tending to be more conservative and discriminating in their assessments of candidate responses. For example, Szandała (2025) found that both LLMs and human experts displayed inconsistency in their judgments, but the nature of their variability differed: LLMs could produce different results for identical inputs upon repeated evaluation, while human experts' variability was influenced by subjective factors such as fatigue or mood. Furthermore, LLMs and humans apply different evaluation philosophies—humans often weigh personal characteristics and contextual nuances more heavily, whereas LLMs focus more on task performance and may be influenced by the structure and content of the input data (Fumagalli et al., 2022).

Additional research highlights that while LLMs can achieve high agreement with human preferences in some contexts, their reliability diminishes in tasks requiring specialized or domain-specific knowledge, where human expertise remains essential (Szymanski et al., 2024). LLMs also tend to show greater score inflation and variability when contextual information is limited, whereas human experts maintain more stable evaluation quality under such conditions (Szandała, 2025). Perceptions of fairness and bias also differ: workers often view human recruiters as more error-prone and subjective, while algorithmic recruiters are seen as more objective but potentially less sensitive to individual or contextual factors (Fumagalli et al., 2022). These findings underscore the importance of retaining human oversight in recruitment, especially for complex or ambiguous cases where nuanced judgment and contextual understanding are critical.

**Contextual Adaptation Mechanisms**

The study identified fundamentally different mechanisms of contextual adaptation between large language models (LLMs) and human experts, highlighting important distinctions in how each processes and responds to environmental information. LLMs, such as GPT, demonstrated statistical sensitivity by adjusting scores and redistributing weights in response to contextual cues like company type. This adaptation was primarily mechanical, relying on the adjustment of predetermined categories rather than a deep understanding of organizational needs or candidate fit, reflecting a more surface-level, data-driven approach to context (Lin et al., 2024; Yang et al., 2025). In contrast, human experts integrated experiential knowledge with specific organizational requirements, offering nuanced insights—such as recognizing the importance of maturity in startup environments versus structured training in multinational corporations—that LLMs could not replicate through quantitative adjustments alone. This qualitative contextual understanding is characteristic of domain expertise and develops through the integration of multiple knowledge sources over time (Gatewood et al., 2015).

The study also found that LLM performance deteriorated significantly under reduced context conditions, supporting concerns about AI systems' dependence on information completeness (Lin et al., 2024). While human experts could draw on professional experience to maintain evaluation quality despite limited information, LLMs lacked compensatory mechanisms for such gaps, leading to increased variability and less reliable outputs. This difference has significant implications for real-world deployment, where the quality and completeness of job descriptions and organizational information can vary widely.

**Implications for Practice**

These findings carry important implications for organizations considering the deployment of large language models (LLMs) in recruitment processes. While LLM-based systems can dramatically increase efficiency—demonstrating screening speeds up to eleven times faster than manual methods and achieving high accuracy in resume classification and grading—systematic differences between LLM and human expert evaluations suggest that direct replacement of human judgment with AI could fundamentally alter selection outcomes (Gan et al., 2024; Vijayalakshmi et al., 2024). Specifically, LLMs may advance different candidate profiles than human experts, potentially misaligning with organizational values or nuanced hiring needs (Köchling & Wehner, 2020).

The reliability of LLMs is highly context-dependent, with performance deteriorating when provided with incomplete or non-standardized information. This underscores the necessity for organizations to invest in comprehensive and standardized job descriptions to ensure consistent and reliable AI-driven screening results (Liu et al., 2021). Without such standardization, brief or preliminary job postings may lead to unreliable or biased outcomes, as LLMs are sensitive to the quality and completeness of contextual information (Gan et al., 2024).

Moreover, the complementary strengths of LLMs and human experts suggest that hybrid approaches may offer optimal solutions. LLMs excel at processing explicit criteria at scale with transparent and auditable decision-making, while human experts contribute contextual understanding, risk assessment, and experiential knowledge that current AI systems cannot replicate (Barocas & Selbst, 2016). Integrating both can leverage the efficiency and consistency of AI while preserving the nuanced judgment essential for fair and effective hiring.

**Theoretical Contributions**

This study advances several theoretical frameworks central to human-AI interaction and decision-making. The results support Signal Detection Theory by showing that reliable signal extraction from both human experts and large language models (LLMs) depends on the quality and completeness of contextual information and the design of the decision system. When context was reduced, the ability of both LLMs and humans to distinguish signal from noise deteriorated, empirically confirming theoretical predictions about the importance of adequate information for reliable decision-making (Szandała, 2025).

The findings also reinforce concerns about the limitations of LLMs in producing consistent and meaningful outputs across varied conditions. LLMs, such as ChatGPT and others, exhibited increased score inflation and inconsistency when faced with ambiguous or incomplete information, suggesting that these models may generate plausible but unreliable outputs under uncertainty. This highlights fundamental architectural limitations in current AI systems, echoing broader critiques about their reliability and interpretability in real-world applications (Dentella et al., 2023; Szymanski et al., 2024).

From the perspective of expertise research, the study demonstrates that human professional expertise involves qualitative understanding and adaptive capabilities that current LLMs cannot fully replicate. Human experts maintained more robust and stable performance under information scarcity, reflecting the kind of transferable, context-sensitive knowledge that characterizes genuine expertise in complex domains (Szandała, 2025; Dentella et al., 2023). In contrast, LLMs' performance was more sensitive to the quality and quantity of input data, underscoring the continued importance of human judgment in high-stakes or ambiguous decision-making contexts.

**Bias and Fairness Implications**

Systematic differences between LLM and human expert evaluations raise critical questions about bias and fairness in algorithmic hiring. Research shows that while LLMs can be designed to promote impartiality and inclusivity, they may still perpetuate or amplify biases present in training data or decision processes, potentially impacting protected groups such as those defined by race, gender, or disability (Gallegos et al., 2023; Tilmes, 2022). For example, studies replicating classic field experiments found that LLMs are generally robust across race and gender, but can display bias related to pregnancy status and political affiliation, and that retrieval models in hiring contexts can exhibit non-uniform selection patterns across demographic groups (Veldanda et al., 2023; Seshadri & Goldfarb-Tarrant, 2025).

A notable concern is the tendency of LLMs to inflate scores under uncertainty, especially when contextual information is limited. This can have disparate impacts if information scarcity is more common among certain candidate groups, such as those with less detailed resumes or job descriptions, potentially leading to systematic disadvantages or advantages (Barocas & Selbst, 2016; Seshadri & Goldfarb-Tarrant, 2025). Additionally, LLMs' varying sensitivity to organizational context may result in different evaluation outcomes for candidates applying to startups versus multinational corporations, which could inadvertently affect career opportunities along lines correlated with protected characteristics (Seshadri & Goldfarb-Tarrant, 2025).

While algorithmic hiring tools offer opportunities to reduce some forms of human bias, they also introduce new risks and complexities. Bias detection and mitigation remain ongoing challenges, requiring careful design, diverse and unbiased training data, and continuous monitoring to ensure equitable outcomes (Gallegos et al., 2023; Koh et al., 2023). Ultimately, a balanced and context-aware approach is necessary to harness the benefits of LLMs in recruitment while minimizing the risk of perpetuating or introducing unfairness.

**Limitations and Future Directions**

Several limitations constrain the generalizability of these findings and highlight important directions for future research. Focusing exclusively on Product Manager positions limits the external validity of the results, as patterns of LLM-human expert alignment may differ across roles with varying skill requirements and evaluation criteria (Cohen, 1988). Future studies should therefore examine a broader range of job roles and organizational contexts to assess whether observed patterns hold more widely (Potočnik et al., 2021).

The specific LLMs evaluated in this study represent current technology, which is evolving rapidly. As such, findings about model capabilities may have limited longevity. However, fundamental questions about distinguishing signal from noise in AI decision-making and the nature of human-AI differences are likely to remain relevant as technology advances (Kaya & Ghosh, 2024). Longitudinal research could help determine whether LLM-human expert differences persist over time and across different candidate populations, providing insight into the stability of these patterns.

The use of an alpha level of 0.1 for primary analyses, while suitable for exploratory research, may increase the risk of Type I errors. Future research should validate these findings using more conservative statistical thresholds to ensure robustness (Benjamini & Hochberg, 1995). Additionally, the reduced human expert sample size due to the loss of one expert in later experimental conditions may have affected the reliability of human-AI comparisons, underscoring the need for adequate sample sizes in comparative studies.

Bias and fairness remain critical priorities for future research, as understanding whether systematic LLM-human differences create disparate impacts on protected groups is essential for responsible AI deployment (Köchling & Wehner, 2020; Hunkenschroer & Luetge, 2022). Research should also explore calibration mechanisms to better align LLM outputs with human expert judgment and investigate ways to improve LLM contextual understanding and resilience to information scarcity (Passerini et al., 2025).

**Conclusion**

This study demonstrates that large language models (LLMs) excel in resume screening but exhibit systematic differences from human experts, posing challenges for direct replacement without significant redesign. LLMs show context-dependent reliability, producing meaningful signals with detailed prompts but noise under information scarcity. GPT displays strong adaptability but risks overcompensation in uncertain conditions, while Claude prioritizes technical consistency with limited contextual flexibility, and Gemini shows selective responsiveness. In contrast, human experts maintain conservative, nuanced evaluations and greater robustness under limited information, leveraging experiential knowledge that LLMs cannot replicate.

These findings highlight opportunities for hybrid recruitment systems, combining LLM efficiency and scalability with human contextual judgment. Organizations should standardize job descriptions to enhance LLM reliability and retain human oversight for complex evaluations. The study advances Signal Detection Theory by linking decision-making reliability to information quality and underscores the unique role of human expertise in high-stakes contexts. These insights guide effective AI integration in recruitment and extend to broader human-AI collaboration, emphasizing the need to balance algorithmic strengths with human judgment for fair and effective outcomes.